# Feature selection strategies for optimized heart disease diagnosis using ML and DL models

Bilal Ahmad[1*] | Jinfu Chen[1*] | Haibo Chen[1]

[1]School of Computer Science and Communication Engineering, Jiangsu University, Zhenjiang 212013, China

Correspondence  *Dr. Bilal Ahmad : bilalrouf@ujs.edu.cn

**Abstract**

Heart disease remains one of the leading causes of morbidity and mortality worldwide, necessitating the development of effective diagnostic tools to enable early diagnosis and clinical decision-making. This study evaluates the impact of feature selection techniques—Mutual Information (MI), Analysis of Variance (ANOVA), and Chi-Square—on the predictive performance of various machine learning (ML) and deep learning (DL) models using a dataset of clinical indicators for heart disease. Eleven ML/DL models were assessed using metrics such as precision, recall, AUC score, F1-score, and accuracy. Results indicate that MI outperformed other methods, particularly for advanced models like neural networks, achieving the highest accuracy of 82.3% and recall score of 0.94. Logistic regression (accuracy 82.1%) and random forest (accuracy 80.99%) also demonstrated improved performance with MI. Simpler models such as Naive Bayes and decision trees achieved comparable results with ANOVA and Chi-Square, yielding accuracies of 76.45% and 75.99%, respectively, making them computationally efficient alternatives. Conversely, k-Nearest Neighbors (k-NN) and Support Vector Machines (SVM) exhibited lower performance, with accuracies ranging between 51.52% and 54.43%, regardless of the feature selection method.

This study provides a comprehensive comparison of feature selection methods for heart disease prediction, demonstrating the critical role of feature selection in optimizing model performance. The results offer practical guidance for selecting appropriate feature selection techniques based on the chosen classification algorithm, contributing to the development of more accurate and efficient diagnostic tools for enhanced clinical decision-making in cardiology.

## 1. Introduction

Cardiovascular disease (CVD) poses a significant global health challenge, accounting for a substantial proportion of deaths worldwide[1]. Early and accurate diagnosis of CVD is crucial for effective treatment and improving patient outcomes. However, the complexity of CVD, arising from multiple risk factors, makes diagnosis challenging. Traditional diagnostic methods often rely on time-consuming procedures and the expertise of specialized physicians, which can limit accessibility and affordability.

Machine learning (ML) has emerged as a powerful tool with the potential to transform CVD diagnosis and prediction. ML algorithms can learn complex patterns from large datasets, enabling the identification of individuals at risk and assisting clinicians in making more informed decisions. The exceptional results achieved by ML and deep learning (DL) in various fields further underscore their potential for healthcare applications. For instance, DL has shown remarkable performance in brain tumor classification by leveraging advanced techniques such as Variational Autoencoders and Generative Adversarial Networks (GANs) [2], [3], as well as in improving skin cancer classification using novel approaches like TED-GAN[4]. Similarly, ML/DL models have been utilized for weakly supervised object localization[5], protein-ligand docking[6], and sentiment analysis during the COVID-19 pandemic[7], [8]. These advancements highlight the versatility and effectiveness of ML/DL approaches in solving complex problems across diverse domains.

In healthcare, ML/DL methods have also excelled in areas like the automatic detection of metastases in stained images[9], predicting cardiovascular events[10]–[12], and integrated diagnosis and segmentation for COVID-19[13]. These successes emphasize the importance of feature selection and model optimization to ensure robust performance. However, a key challenge in developing effective ML models for CVD prediction is the presence of numerous and potentially irrelevant features in healthcare datasets.

Feature selection techniques play a critical role in optimising ML models by identifying and selecting the most relevant features for prediction. These techniques aim to reduce data dimensionality, improve model accuracy, and provide insights into the factors driving CVD. This research focuses on comparing the performance of different feature selection techniques including ANOVA test, chi-square test and mutual information feature selection, in enhancing the accuracy of ML models for heart disease prediction. By evaluating these techniques across a range of ML algorithms and evaluation metrics, this study aims to provide a comprehensive understanding of their impact on CVD prediction model performance. The findings of this research will contribute to the development of more accurate and efficient deep learning-powered tools for CVD diagnosis and risk assessment, ultimately aiding healthcare professionals in delivering timely and effective patient care.

## 2. Related work

Numerous studies have explored the potential of machine learning (ML) for heart disease prediction, employing various algorithms and feature selection techniques. Research in this domain can be broadly categorized into two main areas: studies focusing on optimizing algorithms based on different machine learning techniques and studies investigating the impact of feature selection techniques on model performance. However, there has been limited research directly comparing the effects of diverse feature selection techniques on the performance of heart disease prediction models.

Several studies have focused on comparing the performance of classic machine learning algorithms for heart disease prediction. Premsmith and Ketmaneechairat[14] demonstrated the superiority of logistic regression over neural networks, achieving an accuracy of 91.65% in predicting heart disease. Similarly, Chaurasia and Pal compared Naive Bayes, J48, and bagging algorithms, finding that bagging achieved the highest accuracy (85.03%). Latha and Jeeva investigated ensemble classification techniques, reporting a potential accuracy improvement of up to 7% for weaker classifiers using methods like bagging and boosting. These findings underscore the potential of traditional ML algorithms, but they often lack a systematic evaluation of feature selection—a critical aspect influencing model performance.

More recent research has explored the application of deep learning techniques for heart disease prediction. Mienye and Sun[15] proposed a deep learning strategy using a Particle Swarm Optimization Stacked Sparse Autoencoder (SSAE), achieving an accuracy of 96.1% on the Cleveland dataset. Similarly, Drod et al. (2022)[16] employed ML techniques, including logistic regression and PCA, to identify cardiovascular disease (CVD) risk factors in MAFLD patients. Their analysis of 191 patients revealed hypercholesterolemia, plaque scores, and diabetes duration as key predictors, with their ML model achieving an AUC of 0.87, showcasing its potential for identifying high-risk individuals based on simple clinical data. Al Bataineh and Manacek[17] further contributed by developing a hybrid algorithm combining Multilayer Perceptron (MLP) and Particle Swarm Optimization (PSO), achieving an accuracy of 84.6% and outperforming other traditional algorithms. While these studies highlight the potential of various ML and deep learning algorithms, they often overlook the crucial role of feature selection in optimizing model performance. Addressing this gap, a limited number of studies have investigated the impact of feature selection techniques, primarily focusing on filter methods. Hosam et al.[18] used the chi-square test for feature selection and employed SHAP for explainable AI, demonstrating the importance of feature selection for improving prediction accuracy and explainability. However, the scope of feature selection techniques explored in heart disease prediction research remains narrow, often limited to a small subset of methods.

In addition to algorithmic approaches, Gupta et al.[19] emphasized the importance of clinical and lifestyle factors in heart disease diagnosis and management. They highlighted the use of medical history, physical examination, blood tests, and imaging tests, for diagnosis, alongside lifestyle modifications such as losing weight, managing stress, exercising, and quitting smoking. However, while these insights are valuable, they do not address the methodological advancements required to enhance predictive models.

Chintan et al.[20] proposed a clustering technique with Huang initialization, claiming improved classification performance. Despite these contributions, research on feature selection in heart disease prediction remains fragmented, often focusing on filter methods without exploring the broader range of techniques, such as wrapper and evolutionary methods.

Building on these findings, this study aims to address the gap by comprehensively evaluating the impact of diverse feature selection techniques on heart disease prediction models. Specifically, we applied three feature selection methods: ANOVA F-value, mutual information (MI), and the chi-square test, to analyze their effects on model performance. These techniques were integrated into various ML and DL algorithms to identify the most effective combinations for heart disease prediction. By comparing their performance across multiple evaluation metrics, this research provides valuable insights into developing more accurate and efficient AI-powered tools for CVD diagnosis and risk assessment, thus contributing to the advancement of personalized healthcare solutions.

## 3. Proposed methodology

This study aims to enhance the accuracy of machine learning models in predicting heart disease by utilising feature selection techniques. The proposed methodology involves identifying and selecting the most influential features from the Cleveland Heart Disease dataset. This will be achieved through the application of three distinct feature selection methods: Mutual feature information, Chi-square Test, and ANOVA F-test. By reducing the number of input features, the goal is to improve the efficiency and effectiveness of the prediction models. Figure 1 shows the overview of training ML & DL models.

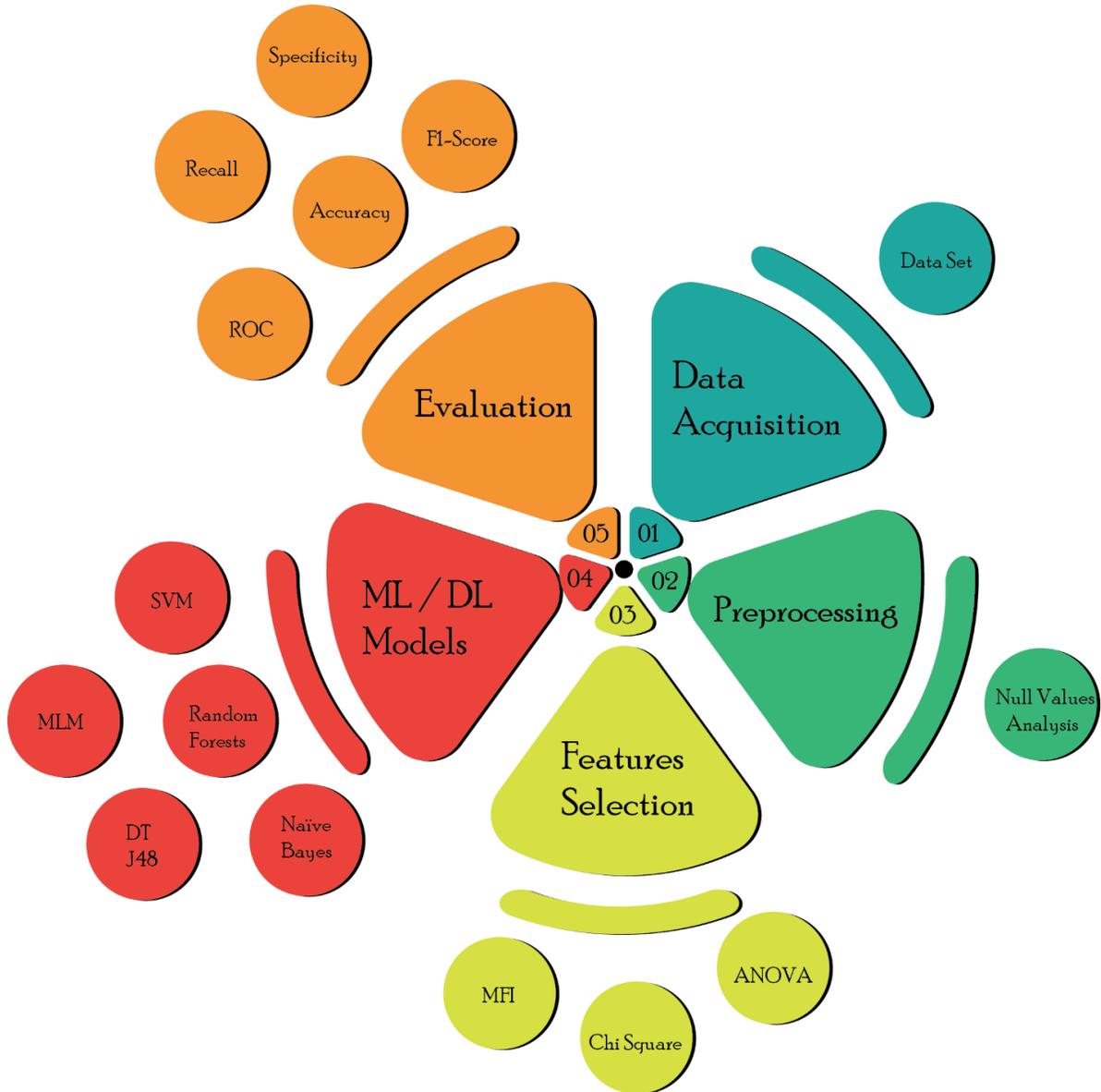

**Figure 1**: Basic workflow of training machine learning and deep learning models for cardiovascular disease prediction.

### 3.1. Dataset

The Cleveland Heart Disease dataset, sourced from the UCI Machine Learning Repository, will be used for this study. This dataset comprises 303 patient records, each containing information on 14 attributes potentially associated with heart disease. These attributes include factors such as age, sex, chest pain type, resting blood pressure, and serum cholesterol levels. Additionally, the dataset includes a target variable indicating the presence or absence of heart disease in each patient.

Table 1: Characteristics of dataset used in this study

| | Feature Code | Feature Name | Description | Data Type |
|---|---|---|---|---|
| 1 | EIA | Exang | Presence of exercise-induced angina. | Binary: 1 = Pain present, 0 = No pain |
| 2 | REC | RestEcg | Resting electrocardiographic results. | Categorical: 0 = No abnormalities, 1 = Normal, 2 = Possible/certain left ventricular hypertrophy |
| 3 | AGE | Age | Patient's age in years. | Numerical (Continuous) |
| 4 | CPT | CP | Type of chest pain experienced by the patient. | Categorical: 1 = Typical angina, 2 = Atypical angina, 3 = Non-anginal pain, 4 = Asymptomatic |
| 5 | FBS | Fbs | Fasting blood sugar level. | Binary: 1 = >120 mg/dl, 0 = <120 mg/dl |
| 6 | MHR | Thalach | Maximum heart rate achieved during exercise. | Numerical (Continuous) |
| 7 | BRP | Trestbps | Resting blood pressure (mm Hg). | Numerical (Continuous) |
| 8 | CHOL | Chol | Patient's cholesterol level (mg/dl). | Numerical (Continuous) |
| 9 | OP | Oldpeak | ST depression induced by exercise relative to rest. | Numerical (Continuous) |
| 10 | CMV | Ca | Number of major blood vessels (0-3) colored by fluoroscopy. | Numerical (Discrete) |
| 11 | PES | Slope | Slope of the peak exercise ST segment. | Categorical: 0 = Upsloping, 1 = Flat, 2 = Downsloping |
| 12 | TS | Thal | Results of the thallium stress test. | Categorical: 0 = Negative, 1 = Positive, 2 = Inconclusive |
| 13 | GEN | Gender | Patient's sex. | Binary: 0 = Female, 1 = Male |
| 14 | Target | Target | Diagnosis of heart disease based on angiographic results. | Binary: 0 = No heart disease (<50% narrowing), 1 = Heart disease (>50% narrowing) |

## 3.2. Features selection techniques

Three distinct feature selection techniques are employed to identify and select the most relevant attributes for predicting heart disease.

I. **ANOVA F-test:** This statistical test is utilised to evaluate the relationship between numerical features and the target variable. The ANOVA F-test help in identifying features that exhibit statistically significant differences in their mean values between patients with and without heart disease. Features displaying such significant differences are selected.
II. **Chi-square Test:** This statistical test is applied to assess the association between categorical features and the target variable. This test helps in determining whether there is a statistically significant relationship between a specific categorical feature and the occurrence of heart disease. Features showing a significant association are selected.
III. **Mutual Feature Information:** This method quantifies the statistical dependency between each feature and the target variable, which represents the presence or absence of heart disease. Features demonstrating a higher mutual information score, indicating a stronger relationship with the target variable, is prioritised for selection.

The implementation details of all three features selection methods ANOVA test, Chi-square test and mutual information, are presented in algorithms 1,2 and 3, respectively.

### *Algorithm 1: Features selection method by using ANOVA test*

*Data: $D$: the dataset with numerical features $X$ and categorical target variable $Y$, $\alpha$: significance level*

*Result: $S$: the set of selected features*

1. $S \leftarrow \emptyset$; // Initialize the set of selected features
2. for each numerical feature $x \in X$ do

3. // Check ANOVA Assumptions

4. **residuals** ← CalculateResiduals($x$, $Y$); // Calculate residuals for feature $x$
5. if **residuals** violate normality then
6.   $x$ ← Transform($x$); // Apply data transformation (e.g., log, sqrt)
7.   **residuals** ← CalculateResiduals($x$, $Y$); // Recalculate residuals after transformation
8. end

9. if **residuals** still violate normality or $x$ violates homoscedasticity (via Levene's test) then
10.   continue; // Skip feature if ANOVA assumptions cannot be met
11. end

12. *// Perform ANOVA Test*

13. **groups** ← GroupData($D$, $x$, $Y$); // Group data by target variable categories
14. **SSB** ← CalculateSSB(**groups**); // Calculate Sum of Squares Between Groups
15. **SSW** ← CalculateSSW(**groups**); // Calculate Sum of Squares Within Groups
16. **dfB** ← NumberOfGroups(**groups**) - 1; // Calculate Between-Groups Degrees of Freedom

17. **dfW** ← TotalObservations($D$) - NumberOfGroups(**groups**); // Calculate Within-Groups Degrees of Freedom
18. **MSB** ← **SSB** / **dfB**; // Calculate Mean Square Between Groups

| |
|---|
| 19. **MSW** ← **SSW** / **dfW**; // Calculate Mean Square Within Groups |
| 20. **F** ← **MSB** / **MSW**; // Calculate F-statistic |
| 21. **p** ← CalculatePValue(**F**, **dfB**, **dfW**); // Calculate p-value from F-distribution |
| **22. // Feature Selection** |
| 23. *if AdjustPValue(**p**, **α**) ≤ **α** then // Apply multiple testing correction (e.g., Bonferroni)* |
| 24.     **S** ← **S** ∪ {**x**}; // Add feature to the selected set |
| 25.   end |
| 26. end |
| 27. return **S**; |

The ANOVA feature selection algorithm was implemented with a significance level (α) of 0.05. Where necessary, data transformations were applied to address violations of normality or homoscedasticity; specifically, a log transformation was applied to 'Chol' and a square root transformation was applied to 'Trestbps'. If transformations did not adequately address these violations, the feature was excluded from consideration. No correction for multiple comparisons was applied. The helper functions Transform(), GroupData(), CalculateSSB(), CalculateSSW(), NumberOfGroups(), TotalObservations() and CalculatePValue() encapsulate the specific implementation details of data transformation, data grouping, sum of squares calculations, number of groups and total number of observations calculations and p-value calculation from F-distribution, respectively.

| *Algorithm 2: Chi-Square feature selection* |
|---|
| *Data: **D**: the dataset with categorical features **X** and categorical target variable **Y**, **α**: significance level* |
| *Result: **S**: the set of selected features* |
| 1. **S** ← ∅; // Initialize the set of selected features |
| 2. *for each categorical feature **x** ∈ **X** do* |
| 3.   // Create Contingency Table |
| 4.   **contingency_table** ← CreateContingencyTable(**D**, **x**, **Y**); |
| 5.   // Calculate Expected Frequencies |
| 6.   **expected_frequencies** ← CalculateExpectedFrequencies(**contingency_table**); |
| 7.   // Calculate Chi-Square Statistic |
| 8.   **χ²** ← CalculateChiSquare(**contingency_table**, **expected_frequencies**); |
| 9.   // Calculate Degrees of Freedom |
| 10.  **df** ← CalculateDegreesOfFreedom(**contingency_table**); |
| 11.  // Calculate p-value |
| 12.  **p** ← CalculatePValueChiSquare(**χ²**, **df**); |
| 13. // Feature Selection |
| 15.     **S** ← **S** ∪ {**x**}; // Add feature to the selected set |
| 16.   end |
| 17. end |
| 18. return **S**; |

The Chi-Square feature selection algorithm was implemented with a significance level ($\alpha$) of 0.05. No correction for multiple comparisons was applied. The algorithm uses the following helper functions, which encapsulate the implementation details and equations:

1. **CreateContingencyTable():** Constructs a contingency table that summarizes the frequencies of combinations between the feature values and the target variable. The table is a two-dimensional matrix where rows represent the unique values of the feature, and columns represent the unique values of the target variable.
2. **CalculateExpectedFrequencies():** Computes the expected frequencies for each cell in the contingency table based on the assumption of independence between the feature and the target variable. The formula for the expected frequency for a cell is:

$$E_{ij} = \frac{R_i \cdot C_j}{N} \qquad (I)$$

where $E_{ij}$ is the expected frequency for cell $(ij)$, $R_i$ is the row total, $C_j$ is the column total, and $N$ is the overall total of the contingency table.

3. **CalculateChiSquare():** Calculates the Chi-Square statistic using the formula:

$$X^2 = \sum \frac{(O_{ij} - E_{ij})^2}{E_{ij}} \qquad (II)$$

where $O_{ij}$ is the observed frequency for cell *(i,j)*, and $E_{ij}$ is the expected frequency for the same cell.

4. **CalculateDegreesOfFreedom():** Determines the degrees of freedom for the contingency table using the formula:

$$df = (\text{Number of Rows} - 1) \times (\text{Number of Columns} - 1) \qquad (III)$$

where the degrees of freedom reflect the number of independent comparisons in the table.

5. **CalculatePValueChiSquare():** Calculates the p-value associated with the Chi-Square statistic using the Chi-Square distribution and the computed degrees of freedom. The p-value indicates the probability of observing the test statistic $X^2$ or something more extreme under the null hypothesis.

These helper functions enable modular implementation of the Chi-Square test, streamlining the processes of contingency table creation, expected frequency computation, Chi-Square statistic calculation, degrees of freedom determination, and p-value computation. This modular design facilitates extendability of the algorithm.

---

*Algorithm 3: Mutual Information (MI) feature selection*

---

*Data: **D**: The dataset with features **X** and target variable **Y**,*
*m: Number of features to select*
*Result: **S**: The set of selected features*

---

*1. **S** ← ∅; // Initialize the set of selected features*
*2. for each feature x ∈ **X** do*

> 3. // Estimate Mutual Information
> 4. $I(x; Y) \leftarrow$ CalculateMutualInformation(x, Y);
> 5. end
> 6. // Rank features by their Mutual Information values
> 7. $\mathbf{R} \leftarrow$ Sort($\mathbf{X}$, descending by $I(x; Y)$);
> 8. // Select top-m features
> 9. $\mathbf{S} \leftarrow$ Top-m features from $\mathbf{R}$;
> 10. return $\mathbf{S}$;

The Mutual Information feature selection algorithm was implemented to identify the top mm most informative features relative to the target variable. Mutual information $(I(x;Y))$ quantifies the dependency between a feature xx and the target variable $Y$, considering both linear and non-linear relationships. For each feature, mutual information was computed using the formula:

$$I(x; Y) = \sum_{x,y} P(x, y) \, \log \frac{P(x,y)}{P(x)P(y)} \qquad (IV)$$

where $P(x, y)$ is the joint probability of feature $x$ and target $Y$, and $P(x)$, $P(y)$ are their respective marginal probabilities. Features were ranked based on their mutual information scores, and the top mm features were selected. The helper function CalculateMutualInformation() encapsulates the computation of mutual information, while the Sort() function ranks features in descending order of mutual information scores. This algorithm performed best among three for identifying both linear and non-linear dependencies between features and the target variable.

### 3.3. Machine learning and deep learning models

Machine Learning (ML) and Deep Learning (DL) are sub branches of artificial intelligence. ML focuses on enabling computers to learn from data without explicit programming, while Deep learning is a subfield of ML that uses artificial neural networks with multiple layers to extract higher-level features from data.
We employed a variety of machine learning models to predict heart disease, both with the original complete set of features and with the reduced feature sets obtained from each of the three feature selection methods. The models are then trained and tested on the data, and their performance is compared to assess the impact of feature selection on their predictive capabilities. The algorithms to be investigated include the following.

**Deep Learning (DL):**

I. **Neural Networks:** Neural networks are a class of DL models inspired by the structure of the human brain. They consist of interconnected layers of nodes (neurons) that process and transform data. They are capable of learning complex non-linear relationships and are particularly effective with large datasets. Different architectures exist, such as Multilayer Perceptrons (MLPs), Convolutional Neural Networks (CNNs), and Recurrent Neural Networks (RNNs).

**Machine Learning (ML):**

II. **Linear Discriminant Analysis (LDA):** LDA is a linear classification method that aims to find a linear combination of features that best separates two or more classes. It assumes that the data within each class is normally distributed and has equal covariance matrices. LDA is computationally efficient and works well for linearly separable data.
III. **Support Vector Machines:** SVMs aim to find the optimal hyperplane that maximally separates data points of different classes. They use kernel functions to map data into higher-dimensional spaces, enabling them to handle non-linearly separable data. Support Vector Classifier (SVC) is the version of SVM for classification tasks.
IV. **Random Forest:** Random forest is an ensemble learning method that constructs multiple decision trees during training and outputs the class that is the mode of the classes (classification) or mean/average prediction (regression) of the individual trees. It reduces overfitting and improves generalization by averaging the predictions of multiple trees.
V. **Gradient Boosting:** Gradient boosting is another ensemble method that builds trees sequentially, with each new tree correcting the errors of the previous trees. It minimizes a loss function using gradient descent. Popular implementations include XGBoost, LightGBM, and CatBoost.
VI. **AdaBoost (Adaptive Boosting):** AdaBoost is an ensemble learning method that combines multiple weak learners (typically decision trees with a single split) to create a strong classifier. It assigns weights to each training instance and focuses on misclassified instances in subsequent iterations.
VII. **Decision Tree:** Decision trees are tree-like structures that recursively partition the data based on feature values. They are easy to interpret and visualize but can be prone to overfitting if not pruned or regularized.
VIII. **Nu SVC:** Nu-Support Vector Classification is a variation of the standard SVC. Instead of the C parameter (which controls the trade-off between maximizing the margin and minimizing classification error), Nu-SVC uses the nu parameter. Nu roughly controls the number of support vectors and thus the complexity of the model. It offers an alternative way to regularize the model.
IX. **k-Nearest Neighbors (k-NN):** k-NN is a simple instance-based learning algorithm that classifies a data point based on the majority class among its k nearest neighbors in the feature space. It's easy to implement but can be computationally expensive for large datasets and sensitive to irrelevant features.
X. **Logistic Regression:** Logistic regression is a linear classification model that uses a sigmoid function to predict the probability of a data point belonging to a particular class. It's widely used for binary classification problems and provides interpretable probability estimates.
XI. **Naive Bayes:** Naive Bayes is a probabilistic classification algorithm based on Bayes' theorem. It assumes that the features are conditionally independent[4] given the class label, which is often a simplifying (and "naive") assumption. It's computationally efficient and performs well in many real-world applications, especially text classification.

### 3.4. Performance measures

To evaluate the effectiveness of the feature selection methods and the performance of the machine learning models, various performance metrics are employed. These metrics help us to determine the impact of feature selection on the predictive accuracy and efficiency of the models. Table 2 contains the performance measures used in this research.

**Table 2:** Various performance measures used in this research

| Accuracy | The proportion of correctly classified instances out of the total number of instances. | $\frac{TP + TN}{(TP + TN + FP + FN)}$ |
|---|---|---|
| Precision | The proportion of true positive predictions out of the total number of positive predictions. | $\frac{TP}{TP + FP}$ |
| Sensitivity/Recall: | The proportion of true positive predictions out of the total number of actual positive instances. | $\frac{TP}{TP + FN}$ |
| Specificity | The proportion of true negative predictions out of the total number of actual negative instances. | $\frac{TN}{TN + FP}$ |
| F1-score | A harmonic mean of precision and recall, providing a balanced measure of the model's performance. | $2 * \left(\frac{Precision * Recall}{Precision + Recal}\right)$ |
| ROC_AUC | The area under the Receiver Operating Characteristic (ROC) curve, which plots the true positive rate against the false positive rate. | - |

## 4. Results and discussions

This study evaluated the influence of feature selection methods on the predictive performance of various machine learning (ML) and deep learning (DL) models for heart disease classification. The feature selection techniques examined were Mutual Information (MI), Analysis of Variance (ANOVA), and Chi-Square, applied to a dataset of clinical indicators. The predictive performance of eleven ML/DL models was assessed using precision, recall, AUC score, F1-score, and accuracy metrics. The results, summarized in Tables 3 through 6, reveal several important insights into the interplay between feature selection techniques and model performance.

Mutual Information feature selection demonstrated the best overall performance for most models. Neural networks achieved the highest accuracy (82.3%) and recall (0.94) among all models and feature selection methods when paired with MI, underscoring the strength of this method in capturing both linear and non-linear dependencies between features and the target variable. Logistic regression and random forest also showed slight performance improvements with MI (Table 3) compared to ANOVA and Chi-Square, achieving accuracies of 82.1% and 80.99%, respectively.

**Table 3:** Heart disease prediction performance of ML/DL models with mutual information features selection

|   | ML/DL Model | Precision | Recall | ROC_AUC | F1 | Accuracy |
|---|---|---|---|---|---|---|
| 1 | Neural Networks | 0.81 | 0.94 | 0.89 | 0.79 | 82.3 |
| 2 | Linear DA | 0.78 | 0.88 | 0.89 | 0.81 | 81.62 |

|    | ML/DL Model | Precision | Recall | ROC_AUC | F1 | Accuracy |
|----|-------------|-----------|--------|---------|-----|----------|
| 3  | Support Vectors | 0.55 | 0.87 | 0.77 | 0.67 | 61.65 |
| 4  | Random Forest | 0.79 | 0.82 | 0.91 | 0.81 | 80.99 |
| 5  | Gradient Boosting | 0.75 | 0.87 | 0.9 | 0.81 | 77.21 |
| 6  | AdaBoost | 0.73 | 0.91 | 0.85 | 0.81 | 78.17 |
| 7  | Decision Tree | 0.78 | 0.81 | 0.8 | 0.79 | 78.31 |
| 8  | Nu SVC | 0.72 | 0.88 | 0.89 | 0.8 | 76.82 |
| 9  | Nearest Neighbors | 0.52 | 0.31 | 0.6 | 0.4 | 52.53 |
| 10 | Logistic Regression | 0.78 | 0.88 | 0.9 | 0.83 | 82.1 |
| 11 | Naive Bayes | 0.76 | 0.84 | 0.91 | 0.81 | 78.17 |

**Table 4:** Heart disease prediction performance of ML/DL models with ANOVA feature selection

|    | ML/DL Model | Precision | Recall | ROC_AUC | F1 | Accuracy |
|----|-------------|-----------|--------|---------|-----|----------|
| 1  | Neural Networks | 0.66 | 0.92 | 0.85 | 0.77 | 75.34 |
| 2  | Linear DA | 0.78 | 0.86 | 0.89 | 0.81 | 81.71 |
| 3  | Support Vectors | 0.55 | 0.85 | 0.76 | 0.67 | 61.94 |
| 4  | Random Forest | 0.78 | 0.8 | 0.89 | 0.79 | 80.14 |
| 5  | Gradient Boosting | 0.75 | 0.85 | 0.88 | 0.79 | 78.83 |
| 6  | AdaBoost | 0.72 | 0.89 | 0.83 | 0.79 | 78.83 |
| 7  | Decision Tree | 0.77 | 0.79 | 0.78 | 0.78 | 78.83 |
| 8  | Nu SVC | 0.7 | 0.87 | 0.88 | 0.78 | 77.52 |
| 9  | Nearest Neighbors | 0.52 | 0.31 | 0.59 | 0.39 | 52.88 |
| 10 | Logistic Regression | 0.8 | 0.87 | 0.89 | 0.82 | 83 |
| 11 | Naive Bayes | 0.76 | 0.82 | 0.9 | 0.78 | 78.83 |

**Table 5:** Heart disease prediction performance of ML/DL models with Chi-Square feature selection

|    | ML/DL Model | Precision | Recall | ROC_AUC | F1 | Accuracy |
|----|-------------|-----------|--------|---------|-----|----------|
| 1  | Neural Networks | 0.65 | 0.9 | 0.82 | 0.75 | 73.04 |
| 2 | Linear DA | 0.76 | 0.84 | 0.87 | 0.8 | 79.43 |
| 3 | Support Vectors | 0.54 | 0.83 | 0.74 | 0.65 | 60.17 |
| 4 | Random Forest | 0.77 | 0.79 | 0.87 | 0.78 | 78.07 |
| 5 | Gradient Boosting | 0.73 | 0.84 | 0.86 | 0.77 | 76.71 |
| 6 | AdaBoost | 0.71 | 0.87 | 0.81 | 0.77 | 76.71 |
| 7 | Decision Tree | 0.76 | 0.77 | 0.76 | 0.76 | 76.71 |
| 8 | Nu SVC | 0.69 | 0.85 | 0.86 | 0.76 | 75.38 |
| 9 | Nearest Neighbors | 0.51 | 0.3 | 0.57 | 0.39 | 51.52 |
| 10 | Logistic Regression | 0.79 | 0.85 | 0.87 | 0.8 | 80.75 |
| 11 | Naive Bayes | 0.74 | 0.8 | 0.88 | 0.76 | 76.71 |

An interesting observation in results was performance consistency across feature selection methods. Certain models exhibited robust performance regardless of the feature selection method used. For instance, logistic regression and random forest maintained high and relatively stable metrics across MI (Table 3), ANOVA (Table 4), and Chi-Square (Table 5) feature selection. This consistency indicates that these models are less sensitive to the specific subset of features selected, making them suitable candidates for scenarios where computational efficiency or simplicity in feature selection is prioritized. Figure 2 compares the confusion matrices of two important models: neural networks, which achieved the best overall performance, and logistic regression, which demonstrated the most consistent performance across the three feature selection methods.

In contrast, the k-Nearest Neighbors and SVM models consistently underperformed across all feature selection methods. The k-NN model, in particular, exhibited the lowest accuracy (52.53% with MI, 52.88% with ANOVA, and 51.52% with Chi-Square) and poor precision-recall trade-offs. This suggests that these models are highly sensitive to irrelevant or redundant features and may not be suitable for this dataset's characteristics without further preprocessing or dimensionality reduction.

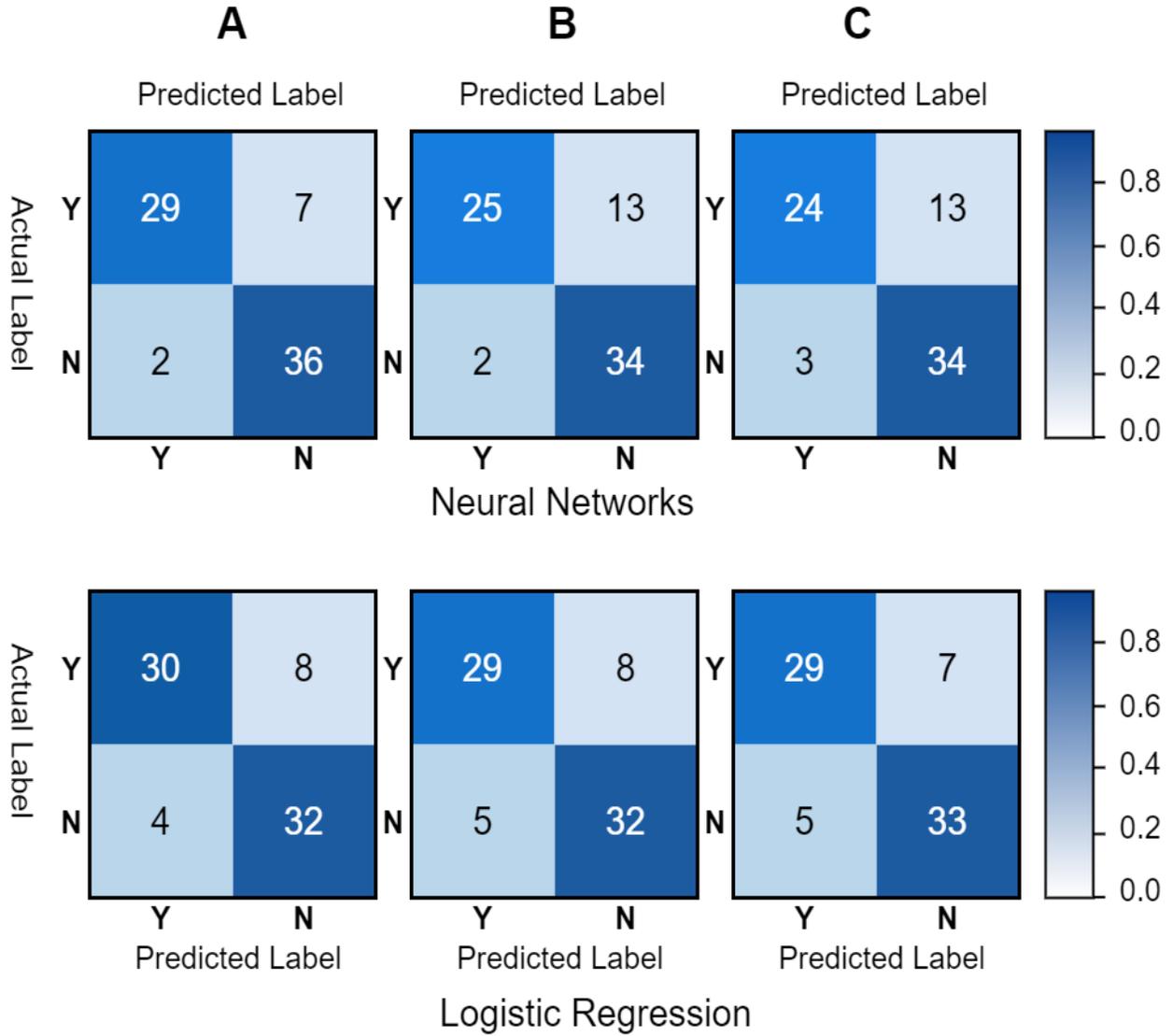

**Figure 2:** Presents the confusion matrices for two representative models: Neural networks (top row), chosen for its peak performance, and logistic regression (bottom row), selected for its stability across different feature selection strategies. The columns represent the feature selection method used: (A) Mutual Information, (B) ANOVA test, and (C) Chi-square test.

**Table 6:** Evaluating the effectiveness of feature selection methods (ANOVA, Chi-Square, Mutual Information) for heart disease classification using ML/DL models

| | ML/DL Model | Precision | | | Recall | | | Accuracy | | |
|---|---|---|---|---|---|---|---|---|---|---|
| | | **ANOVA** | **Chi-Squ.** | **MFS** | **ANOVA** | **Chi-Squ.** | **MFS** | **ANOVA** | **Chi-Squ.** | **MFS** |
| 1 | Neural Networks | 0.66 | 0.65 | 0.81 | 0.92 | 0.94 | 0.94 | 75.34 | 73.04 | 82.3 |
| 2 | Linear DA | 0.78 | 0.76 | 0.78 | 0.86 | 0.84 | 0.88 | 81.71 | 79.43 | 81.62 |
| 3 | SVM | 0.55 | 0.54 | 0.55 | 0.85 | 0.83 | 0.87 | 61.94 | 60.17 | 61.65 |
| 4 | Random Forest | 0.78 | 0.77 | 0.79 | 0.8 | 0.79 | 0.82 | 80.14 | 78.07 | 80.99 |
| 5 | Gradient Boosting | 0.75 | 0.73 | 0.75 | 0.85 | 0.84 | 0.87 | 78.83 | 76.71 | 77.21 |
| 6 | AdaBoost | 0.72 | 0.71 | 0.73 | 0.89 | 0.87 | 0.91 | 78.83 | 76.71 | 78.17 |
| 7 | Decision Tree | 0.77 | 0.76 | 0.78 | 0.79 | 0.77 | 0.81 | 78.83 | 76.71 | 78.31 |
| 8 | Nu SVC | 0.7 | 0.69 | 0.72 | 0.87 | 0.85 | 0.88 | 77.52 | 75.38 | 76.82 |
| 9 | Nearest Neighbors | 0.52 | 0.51 | 0.52 | 0.31 | 0.3 | 0.31 | 52.88 | 51.52 | 52.53 |
| 10 | Logistic Regression | 0.8 | 0.79 | 0.8 | 0.87 | 0.85 | 0.88 | 83 | 80.75 | 82.1 |
| 11 | Naive Bayes | 0.76 | 0.74 | 0.76 | 0.82 | 0.8 | 0.84 | 78.83 | 76.71 | 78.17 |

Where, Chi-Squ. and MFS represents the chi-square test and mutual information features selection.

When comparing the feature selection methods directly (Table 6), MI emerged as the most effective method overall, particularly for complex models like Neural Networks. However, the performance differences between MI, ANOVA, and Chi-Square were not uniform across all models. For simpler models like Naive Bayes and Decision Trees, ANOVA and Chi-Square provided comparable results, with minor differences in precision and recall. This highlights the utility of these simpler techniques in scenarios where computational cost is a concern.

Neural networks displayed notable variability in performance depending on the feature selection method. While MI (Table 3) yielded the best results, ANOVA (Table 4) and Chi-Square (Table 5) resulted in lower accuracy (75.34% and 73.04%, respectively) and precision (0.66 and 0.65). These findings highlight the importance of careful feature selection when working with Neural Networks, as their performance can be significantly enhanced by identifying the most relevant features. Performance comparison of three features selection methods across various machine learning and deep learning models for precision, recall and accuracy is shown in Figure 3.

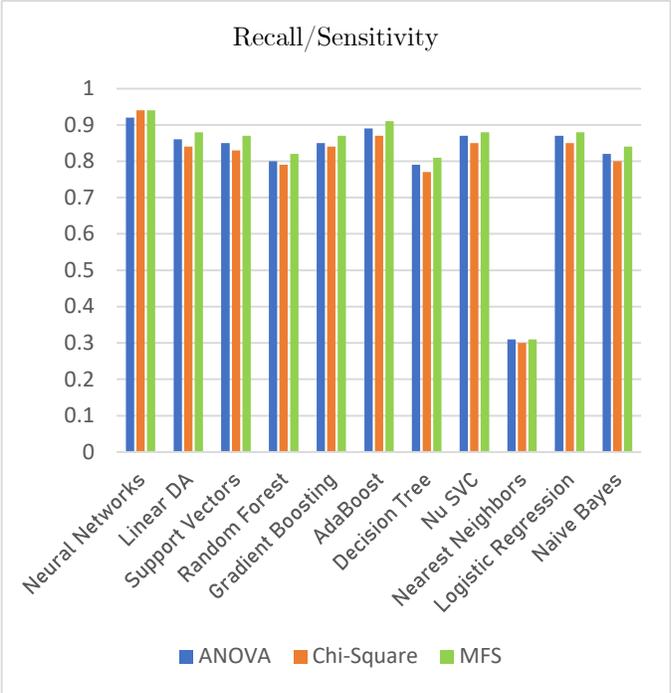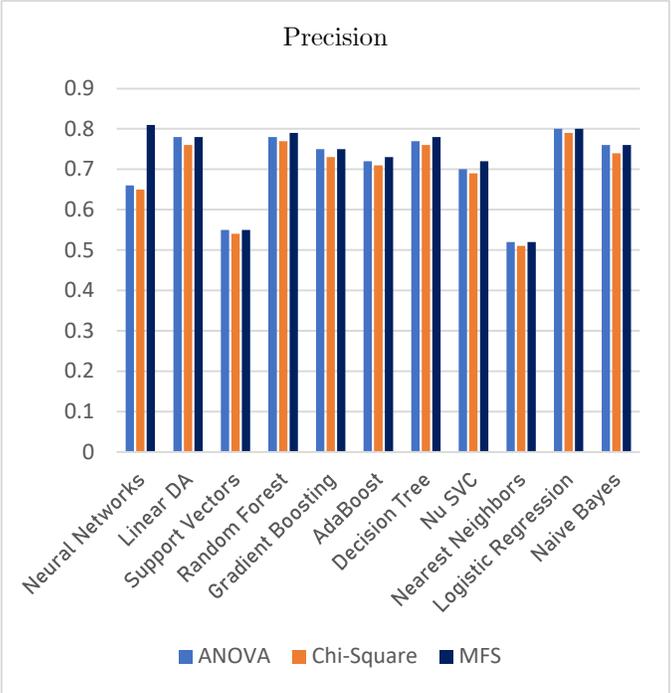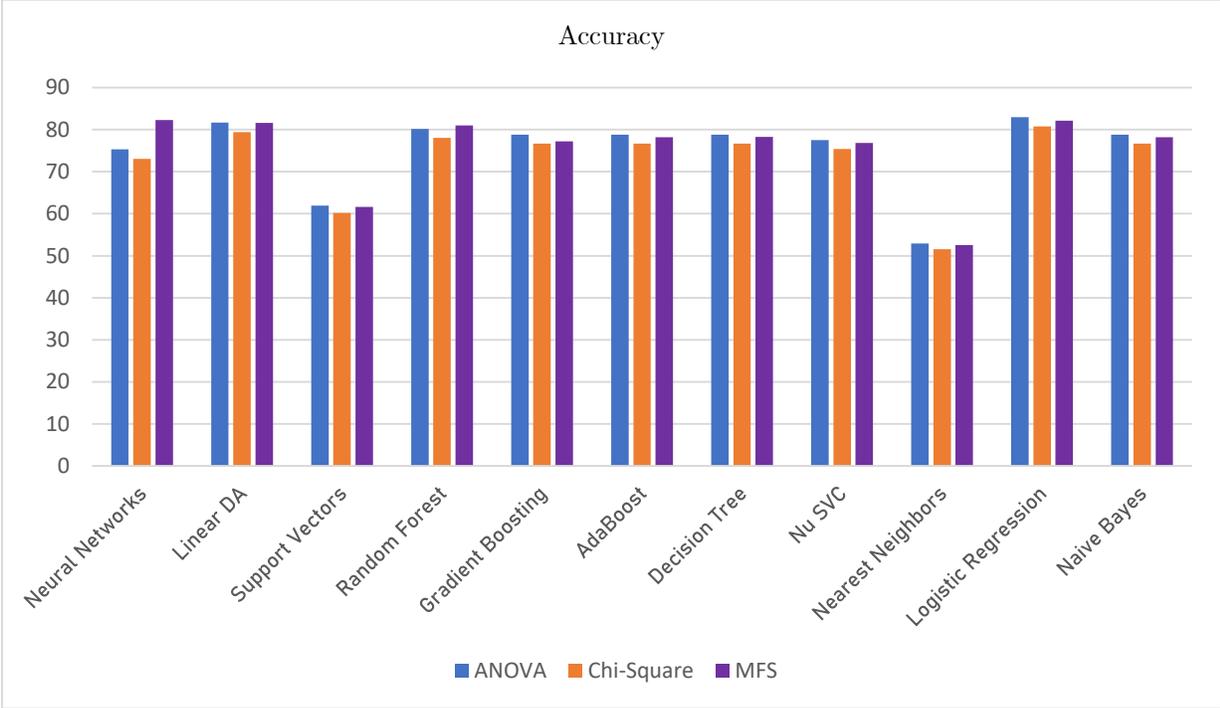

**Figure 3:** Influence of feature selection techniques (ANOVA, Chi-Square, Mutual Information) on various ML & DL models' sensitivity, precision, and accuracy.

This study highlights the critical role of feature selection as a pivotal preprocessing step in enhancing the predictive performance of machine learning and deep learning models for heart disease diagnosis. The findings demonstrate that the choice of feature selection method can substantially influence model outcomes, especially for complex architectures like Neural networks, which showed marked improvements when paired with Mutual Information (MI). This method is recommended for scenarios requiring high accuracy and recall, making it particularly valuable for clinically sensitive applications such as heart disease prediction. On the other hand, for settings where computational efficiency is a priority, simpler methods like ANOVA and Chi-Square offer effective alternatives, performing well with models such as Decision trees and Naive Bayes while maintaining acceptable accuracy levels. However, models like k-NNs and SVM exhibited suboptimal performance, emphasizing the need for additional preprocessing steps such as feature scaling or dimensionality reduction to enhance their robustness.

By providing evidence-based recommendations for tailoring feature selection techniques to specific models and datasets, this study contributes to advancing the field of computational medicine. The insights gained can aid in the development of more accurate and efficient diagnostic tools for heart disease, ultimately supporting clinicians in making well-informed decisions and improving patient outcomes. These findings underscore the importance of interdisciplinary efforts that leverage the power of ML/DL technologies to address pressing challenges in modern medicine.

## 4.1. Limitations and future work

This study's findings are based on a single dataset, which may limit the generalizability of the observed trends. Future work should explore the application of these feature selection techniques and ML/DL models on larger and more diverse datasets to validate the results. Additionally, incorporating advanced feature selection methods, such as Recursive Feature Elimination (RFE) or Embedded Methods, and optimizing hyperparameters for each model could further enhance performance and provide deeper insights into the interplay between feature selection and predictive modeling.

## 5. Conclusion

This study highlights the pivotal role of feature selection in optimizing ML and DL model performance for heart disease prediction. The results and discussion emphasize the critical role of feature selection in improving the accuracy and reliability of machine learning and deep learning models for heart disease prediction, a domain of significant clinical relevance. Among the methods evaluated, Mutual Information (MI) was the most effective, particularly for advanced models like neural networks, which achieved the highest accuracy and recall metrics when paired with MI. Simpler models such as Naive Bayes and Decision Trees performed comparably well with ANOVA and Chi-Square, offering viable, computationally efficient alternatives for clinical applications. In contrast, k-Nearest Neighbors (k-NN) and Support Vector Machines (SVM) showed suboptimal performance with all feature selection methods, highlighting the need for additional preprocessing or dimensionality reduction steps in such cases. These findings underline the importance of selecting feature selection methods that align with the unique characteristics of the dataset and the specific clinical model to enhance prediction outcomes, ultimately contributing to better-informed medical decision-making.